\documentclass[10pt,twocolumn,letterpaper]{article}

\usepackage[algorithms]{wacv}   
\usepackage{wacv}  
\usepackage{graphicx}
\usepackage{amsmath}
\usepackage{amssymb}
\usepackage{booktabs}
\usepackage{subcaption}
\usepackage[normalem]{ulem}
\usepackage[pagebackref,breaklinks,colorlinks]{hyperref}

\usepackage[capitalize]{cleveref}
\crefname{section}{Sec.}{Secs.}
\Crefname{section}{Section}{Sections}
\Crefname{table}{Table}{Tables}
\crefname{table}{Tab.}{Tabs.}

\def\wacvPaperID{1393} 
\def\confName{WACV}
\def\confYear{2025}
\def\AC#1{{\color{black} {{#1}}}} 
\renewcommand{\v}{\mathbf{v}}

\newcommand{\x}{\mathbf{x}}

\newcommand{\z}{\mathbf{z}}

\newcommand{\mA}{\mathtt{A}}

\newcommand{\mD}{\mathtt{D}}

\newcommand{\mI}{\mathtt{I}}

\newcommand{\mV}{\mathtt{V}}

\newcommand{\mX}{\mathtt{X}}

\newcommand{\mZ}{\mathtt{Z}}

\newcommand{\argmin}{\mathop{\mathrm{argmin}}}

\newcommand{\Real}{\mathbb{R}}

\newcommand{\by}{{\times}} 

\begin{document}

\title{Continual Learning in 3D Point Clouds: Employing Spectral Techniques for Exemplar Selection}

\author{Hossein Resani, Behrooz Nasihatkon, Mohammadreza Alimoradi Jazi \\
K. N. Toosi University of Technology\\
\vspace{1em}
{\tt\small hossein.resani@gmail.com} \hspace{1em} 
{\tt\small nasihatkon@kntu.ac.ir} \hspace{1em} 
{\tt\small moradi2reza89@gmail.com}
}
\maketitle

\begin{abstract}
We introduce a novel framework for \underline{C}ontinual \underline{L}earning in \underline{3D} object classification. Our approach, CL3D, is based on the selection of prototypes from each class using spectral clustering. For non-Euclidean data such as point clouds, spectral clustering can be employed as long as one can define a distance measure between pairs of samples. Choosing the appropriate distance measure enables us to leverage 3D geometric characteristics to identify representative prototypes for each class. We explore the effectiveness of clustering in the input space (3D points), local feature space (1024-dimensional points), and global feature space. We conduct experiments on the ModelNet40, ShapeNet, and ScanNet datasets, achieving state-of-the-art accuracy exclusively through the use of input space features. By leveraging the combined input, local, and global features, we have improved the state-of-the-art on ModelNet and ShapeNet, utilizing nearly half the memory used by competing approaches. For the challenging ScanNet dataset, our method enhances accuracy by 4.1\% while consuming just 28\% of the memory used by our competitors, demonstrating the scalability of our approach.\footnote{Project page:  \url{https://doollakh.github.io/cl3d/}}
\end{abstract}


\begin{figure}[t!]
\centering
\begin{subfigure}{.23\textwidth}
  \centering
  \includegraphics[width=.99\linewidth]{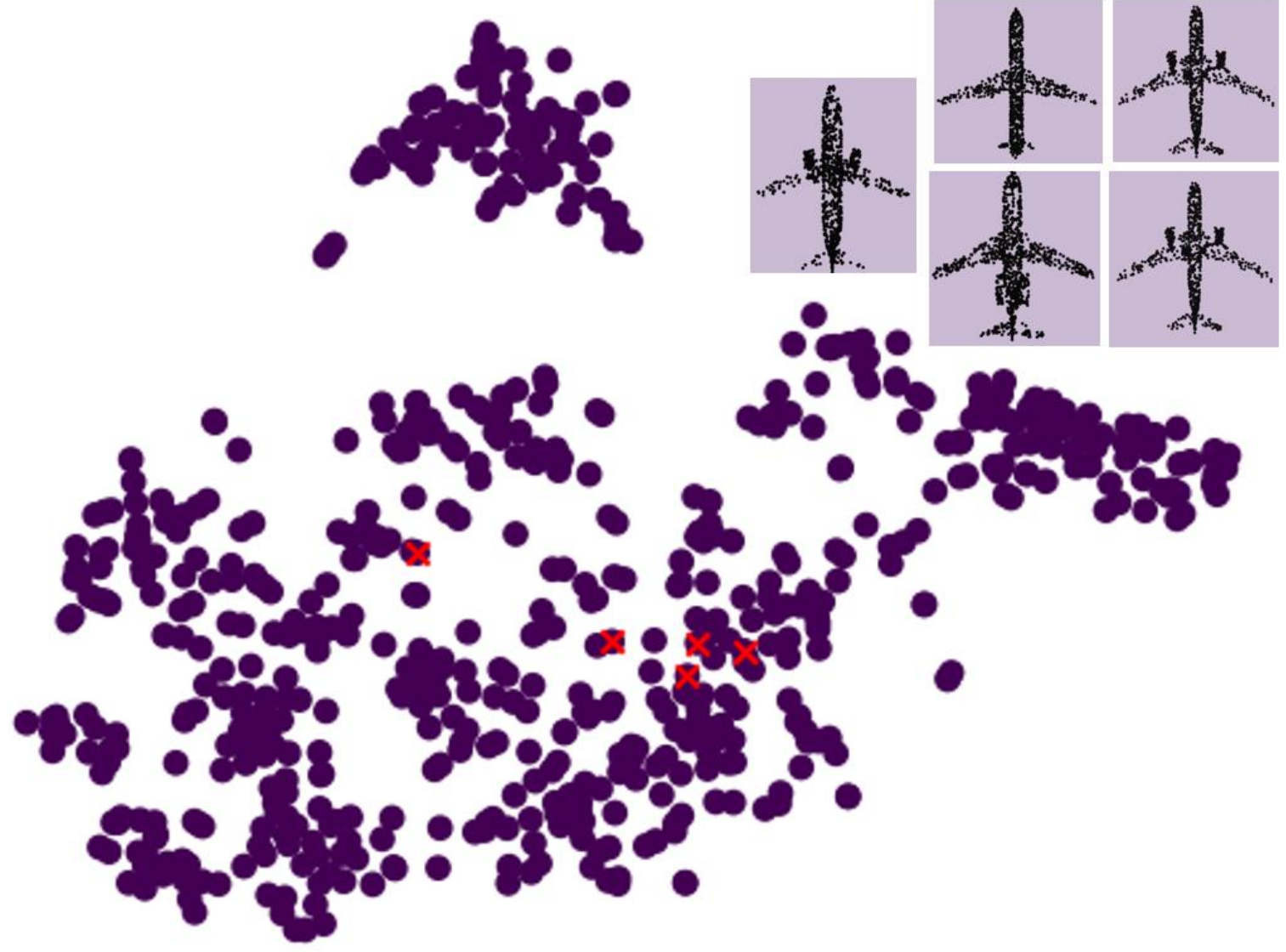}
  \caption*{Herding}
\end{subfigure}%
\hspace{0.01\textwidth}  
\begin{subfigure}{.23\textwidth}
  \centering
  \includegraphics[width=.99\linewidth]{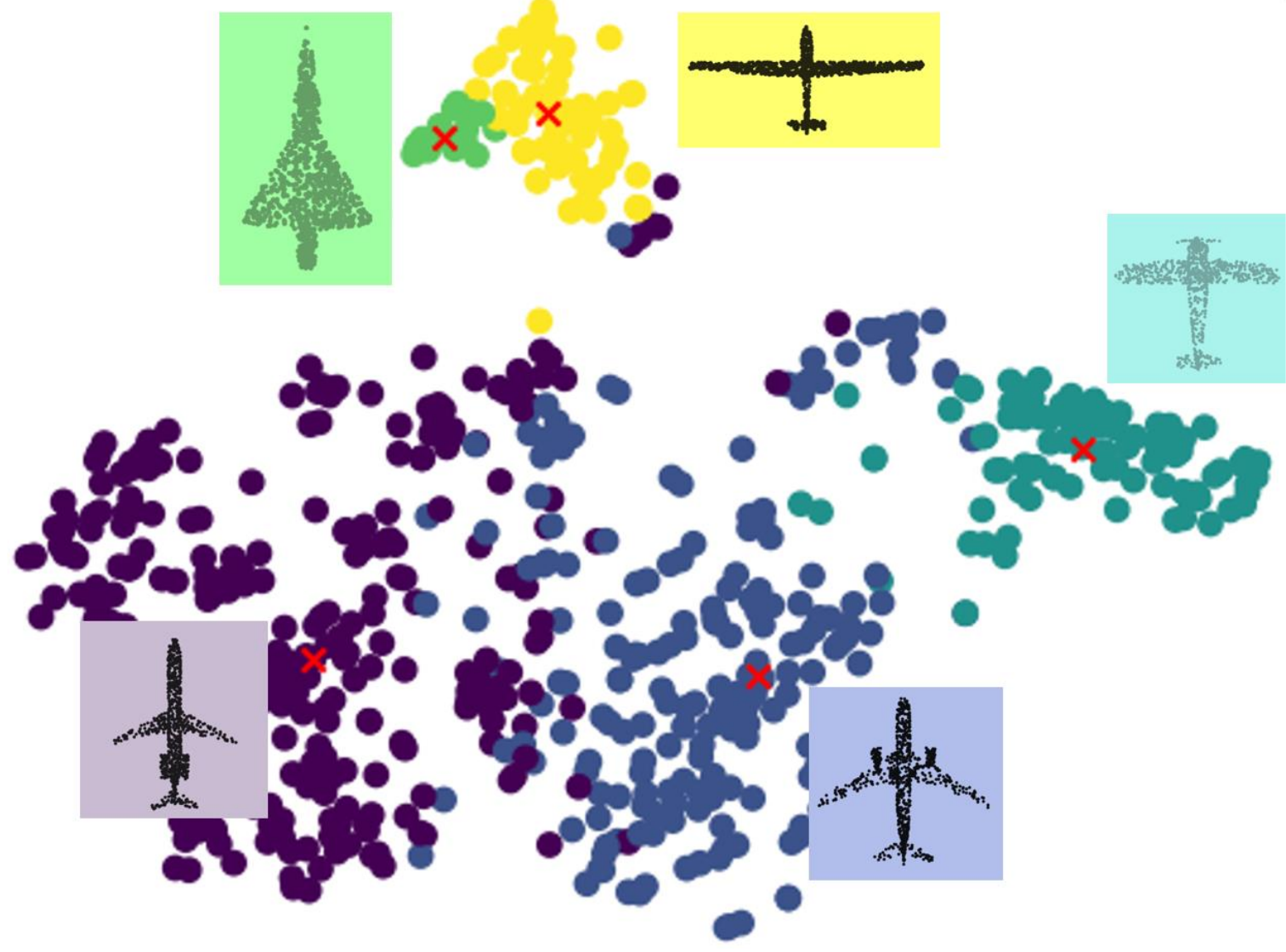}
  \caption*{CL3D (Ours)}
\end{subfigure}
   \caption{t-SNE visualization of global features of the airplane class and exemplars selected by the \emph{herding} approach \cite{Rebuffi_2017_CVPR} (\textbf{left}) contrasted with those from our CL3D method (\textbf{right}). Different colors denote different clusters. Our method effectively covers different subcategories of the airplane samples, demonstrating improved exemplar selection compared to the herding approach.}

   \label{fig:tsne_plane_cluster}
\end{figure}

\section{Introduction} 
\label{sec:intro}

\AC{In recent years, deep learning for point cloud processing has become a key focus in computer vision research due to its wide range of applications\cite{guo2020deep,xiao2024survey}. However, much of the progress has been made under idealized conditions. In real-world applications, gathering data for all object classes simultaneously is often impractical. Typically, models are initially trained on a large dataset for specific classes, known as the base task, to develop a baseline model. As new data for additional classes—referred to as novel tasks—becomes available, the model needs to be updated incrementally. However, retraining the model with both base and novel data is often infeasible due to hardware limitations or privacy concerns. This leads to the problem of \textit{catastrophic forgetting}\cite{kemker2018forgetting}, where the model loses its ability to recognize previously learned classes when adapting to new ones. Continual learning offers a solution by enabling models to learn from new data while retaining knowledge of prior tasks, which is critical for adaptive applications. While continual learning has been extensively studied for 2D images, its application to 3D point cloud data remains largely underexplored.}

3D data representation poses significant challenges compared to 2D due to the irregular structure of point clouds, which lack the uniformity of pixel grids in images. This makes it difficult to apply traditional CNNs, designed for structured data like pixel grids. Instead, specialized techniques such as PointNet\cite{qi2016pointnet}, PointNet++\cite{NIPS2017_pointNet++}, and graph neural networks (GNNs)\cite{shi2020pointgnn,wang2019_DGCNN,yang2018foldingnet} are used to handle the non-Euclidean nature of 3D data. Additionally, the 3D domain suffers from resource scarcity and limited datasets, with ModelNet40\cite{wu20153d} being significantly smaller and less diverse than 2D datasets like ImageNet\cite{deng2009imagenet}. This limitation hampers the ability of 3D models to learn robust and distinct features, complicating classification tasks.\cite{cheraghian2020transductive,ye2023closer,guo2020deep}

The unique characteristics of 3D point clouds directly affect the performance
of memory-based continual learning methods, where a few
exemplars from old classes are stored to aid future learning stages. Traditional exemplar selection methods, such
as \textit{herding} \cite{Rebuffi_2017_CVPR}, become ineffective due to the irregular and
multimodal nature of 3D feature spaces. Moreover, exemplar selection must consider the geometry and structural
properties specific to point clouds to be effective. \cref{fig:tsne_plane_cluster}
provides an illustration of this phenomenon.

To address these challenges, we propose
a novel method for exemplar selection that leverages the geometry of
input point clouds through spectral clustering, which is well-suited
for non-Euclidean data. Additionally, we extend this approach to
perform clustering on local features and employ k-means clustering for
global features. Our contributions include:

\begin{itemize}

\item \textbf{Clustering-based Exemplar Selection:} We demonstrate that clustering for exemplar selection in 3D point clouds, whether applied to input point clouds, local features, or global features, is much more effective than traditional approaches such as \emph{herding} \cite{Rebuffi_2017_CVPR}. 
  
\item \textbf{Geometry-Aware Exemplar Selection:} We introduce an exemplar
  selection method based on spectral clustering that utilizes the
  intrinsic geometric properties of 3D point clouds, independent of
  the backbone architecture. We further extend this to perform
  clustering on local features.

\item\textbf{Fusing Clustering Domains:} We present an innovative approach to fuse
  embeddings from the input space, local feature space, and global
  feature space to enhance exemplar selection via clustering.

\item \textbf{State-of-the-Art Performance:} 
  Our method achieves state-of-the-art results on three prominent
  point cloud benchmarks: ModelNet40\cite{wu20153d},
  ShapeNet\cite{chang2015shapenet}, and
  ScanNet\cite{dai2017scannet}. Averaged over incremental stages, our
  method surpasses previous state-of-the-art methods by 3.4\%, 2\%,
  and 4.1\% on ModelNet40, ShapeNet, and ScanNet, respectively, while
  requiring fewer exemplars in memory. Notably, in the final
  incremental stage, these improvements are astonishing, reaching
  16.9\%, 3.0\%, and 14.0\% on ModelNet40, ShapeNet, and ScanNet,
  respectively.
  
\end{itemize}


\section{Related Work}
 \subsection{Point Cloud Classification}
PointNet \cite{qi2016pointnet} utilizes max-pooling to provide permutation-invariant features suitable for classification tasks. Subsequent studies \cite{NIPS2017_pointNet++,wang2019_DGCNN, GDANet_2021,Walk_2021_ICCV,qiu2021geometric} have developed various architectures to compile information from points nearby, although many continue to use max-pooling to achieve permutation invariance. These techniques are collectively known as point-based methods. In contrast, some techniques \cite{goyal2021revisiting, lawin2017deep} transform 3D point clouds into 2D image representations, applying traditional image processing techniques for predictions.

 \subsection{Continual Learning on 2D Images}
 The issue of catastrophic forgetting \cite{MCCLOSKEY_forget} has been
 a focal point of research. Existing literature typically falls into
 one of three categories: memory-based
 \cite{Rebuffi_2017_CVPR,shin2017continual,LUCIR_2019_CVPR,wu2018memory,Ostapenko_2019_CVPR,EEIL_2018_ECCV},
 regularization-based
 \cite{Chaudhry_2018_ECCV,li2017learning,LWM_2019_CVPR,zenke2017continual},
 and parameter isolation-based strategies
 \cite{Mallya_2018_CVPR,Mallya_2018_ECCV,rusu2016progressive}. Memory-based approaches maintain or recreate exemplars from past
 tasks to integrate with new task training or generate
 new examples for this purpose. Parameter isolation techniques
 dedicate certain parameters exclusively to each task to minimize
 forgetting. Regularization strategies are divided into those that are
 prior-focused, which view knowledge as the values of parameters and
 restrict new task learning by penalizing significant alterations to
 parameters critical for previous tasks
 \cite{zenke2017continual,Chaudhry_2018_ECCV,Aljundi_2018_ECCV}, and
 those that are data-focused, which leverage knowledge distillation
 \cite{hinton2015distilling} by applying a regularization term based
 on the discrepancy between activations of the old and new networks to
 mitigate forgetting \cite{li2017learning,LWM_2019_CVPR}.

 \subsection{Continual Learning on 3D Point Cloud}
While continual learning has seen significant progress in 2D, the 3D domain remains underdeveloped. Chowdhury \textit{et al.} \cite{chowdhury2021learning} used knowledge distillation and semantic word vectors to reduce catastrophic forgetting. Zhao \textit{et al.} \cite{zhao2022static_detection} addressed this issue in 3D object detection with static and dynamic teachers. Zamorski \textit{et al.} \cite{zamorski2023compressed_Rehearsal} introduced Random Compression Rehearsal (RCR), using a compact model to compress and store key data from previous tasks.

\AC{Several studies align with our work. Chowdhury et al. \cite{fscil3d2022} propose using Microshapes—orthogonal basis vectors to represent 3D objects, which help bridge the gap between synthetic and real data, improving model robustness against noise. I3DOL \cite{i3dol2020} addresses irregular point cloud data with an adaptive-geometric centroid module and a geometric-aware attention mechanism to focus on key local structures and reduce forgetting. It also introduces a score fairness compensation strategy to balance training between new and old classes. InOR-Net \cite{inornet2023} enhances this approach with \textit{category-guided geometric reasoning} and \textit{critic-induced geometric attention} to identify key 3D features. It also introduces a dual adaptive fairness compensation strategy to address class imbalances and avoid biased predictions. While these methods rely on the herding algorithm designed for 2D image data, we employ a different exemplar selection strategy that is more effective for 3D point clouds.}

\section{Background}
\subsection{Class-Incremental Learning}
Consider a sequence of disjoint tasks $\mathcal{D} = \{\mathcal{D}^1, ..., \mathcal{D}^T\}$ where the $t$-th
task $\mathcal{D}^t = \{({\mX_i^t, y_i^t})\}_{i=1}^{N_t}$ consists of $N_t$ point clouds samples $\mX_i^t$ and their corresponding class labels $y_i^t \in \mathcal{C}^t$. The classes between all tasks are disjoint, that is $\mathcal{C}^t \cap \mathcal{C}^s = \emptyset$ for $t \neq s$. The goal of continual learning is to progressively train a model, where at each stage $t$, only the training samples $\mathcal{D}^t$ from the current task are accessible. During testing, the model trained on task $\mathcal{D}^t$ is expected to predict outputs not only for the current task but also for all prior tasks $\mathcal{D}^1, ..., \mathcal{D}^{t-1}$.

\subsection{Memory-based Continual Learning}
In \emph{memory-based} continual learning, we maintain a limited memory $\mathcal{M}^{t-1}$ of \emph{exemplars}, which are selected samples from previous tasks $\mathcal{D}^1, \ldots, \mathcal{D}^{t-1}$ to represent old classes. The combined data from the memory and the current task, $\mathcal{M}^{t-1} \cup \mathcal{D}^t$, is used to train the network.

To update $\mathcal{M}^{t}$ after each task, a commonly used strategy for selecting representative exemplars is the \emph{herding} algorithm \cite{Rebuffi_2017_CVPR}. Herding computes class centroids in the feature space and selects exemplars based on their proximity to these centroids. The algorithm iteratively builds $\mathcal{M}^{t}$ by selecting samples closest to each centroid. While herding is widely employed in 2D domains, it faces limitations in 3D object domains due to the multi-modal distribution of each class in feature space \cite{chowdhury2021learning}. In the case of 3D point clouds, herding often selects exemplars from a dominant mode, thereby overlooking the diversity within the data. By contrast, our method captures the complexity and variability of the data by selecting exemplars from multiple modes.

\subsection{PointNet}
\label{sec:pointnet}
In this paper, we present our methodology within the framework of PointNet \cite{qi2016pointnet}. Similar adaptations may be made to other architectures.
Let the $i$-th training sample be $(\mX_i, y_i)$ where $\mX_i \in \Real^{n_i \by 3}$ is the $i$-th input point cloud and $y_i \in \{1,\ldots, C\}$ is the corresponding class label. 
Each row of $\mX_i$ is a single 3D point in the point cloud. In a standard PointNet approach, each input point cloud $\mX_i$ is processed as follows
\begin{enumerate}
    \item First, a shared MLP mapping $f_\theta\colon \Real^3 \rightarrow \Real^F$ is applied to each row of $\mX_i$, resulting in a matrix $\mZ_i \in \Real^{n_i \by F}$. 
    The rows of $\mZ_i$ can be considered as features local to each 3D point in the point cloud. Hence, they are referred to as \emph{local features}. Notice that $\mZ_i \in \Real^{n_i \by F}$ can be viewed as a point cloud in $\Real^F$.

    \item Next, a pooling operator (usually max-pooling) is applied to the columns of $\mZ_i \in \Real^{n_i \by F}$ giving a  \emph{global feature} $\z_i \in \Real^F$. 

    \item Another MLP $g_\phi\colon \Real^F \rightarrow \Real^C$ maps $\z_i$ to $C$ classes.       
\end{enumerate}
The network parameters $(\theta, \phi)$ are learned via backpropagation using a standard cross-entropy loss.

\section{Proposed Method}
\label{sec:proposed}
\begin{figure*}[h]
  \centering

       \includegraphics[width=0.99\linewidth]{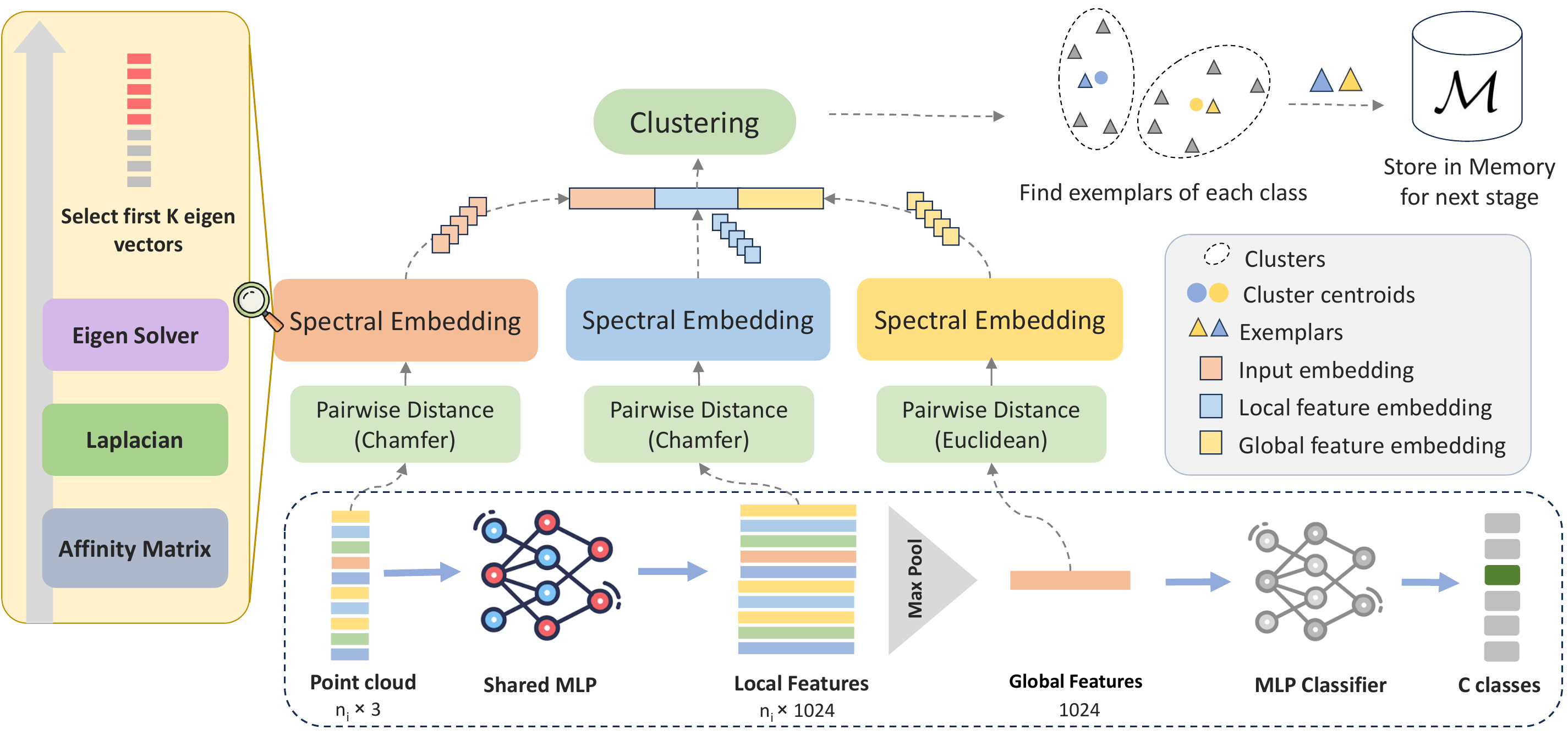}

       \caption{Overview of CL3D training pipeline. Spectral
         embeddings may be computed from the input, the local
         features, and the global features. They can be simply fused
         together by concatenation and fed to $k$-means for clustering.
         Finally, the class exemplars are selected based on 
         proximity to each cluster's centroid. }
    \label{fig:our_model}
\end{figure*}

To perform memory-based continual learning, we choose $K$ candidate exemplars from each old class to store in the memory. 
To do this, we divide the samples of each class into $K$ clusters and choose a single exemplar per cluster. 
The clustering may be applied to the 3D input point clouds $\mX_i$, the local features $\mZ_i$, or the global features $\z_i$. 
Due to their non-Euclidean nature, we perform a spectral embedding on $\mX_i$-s or $\mZ_i$-s before clustering. The global features $\z_i$ could 
be clustered with or without a spectral embedding. We also demonstrate that clustering is most effective when all these three domains are fused together. 
Our pipeline has been depicted in \cref{fig:our_model} and is detailed in the rest of this section. 

\subsection{Spectral Clustering on Point Clouds}
\label{sec:spectral_clustering}
\begin{figure}[t!]
\centering
\begin{subfigure}{.4\textwidth}
  \centering
  \includegraphics[width=.9\linewidth]{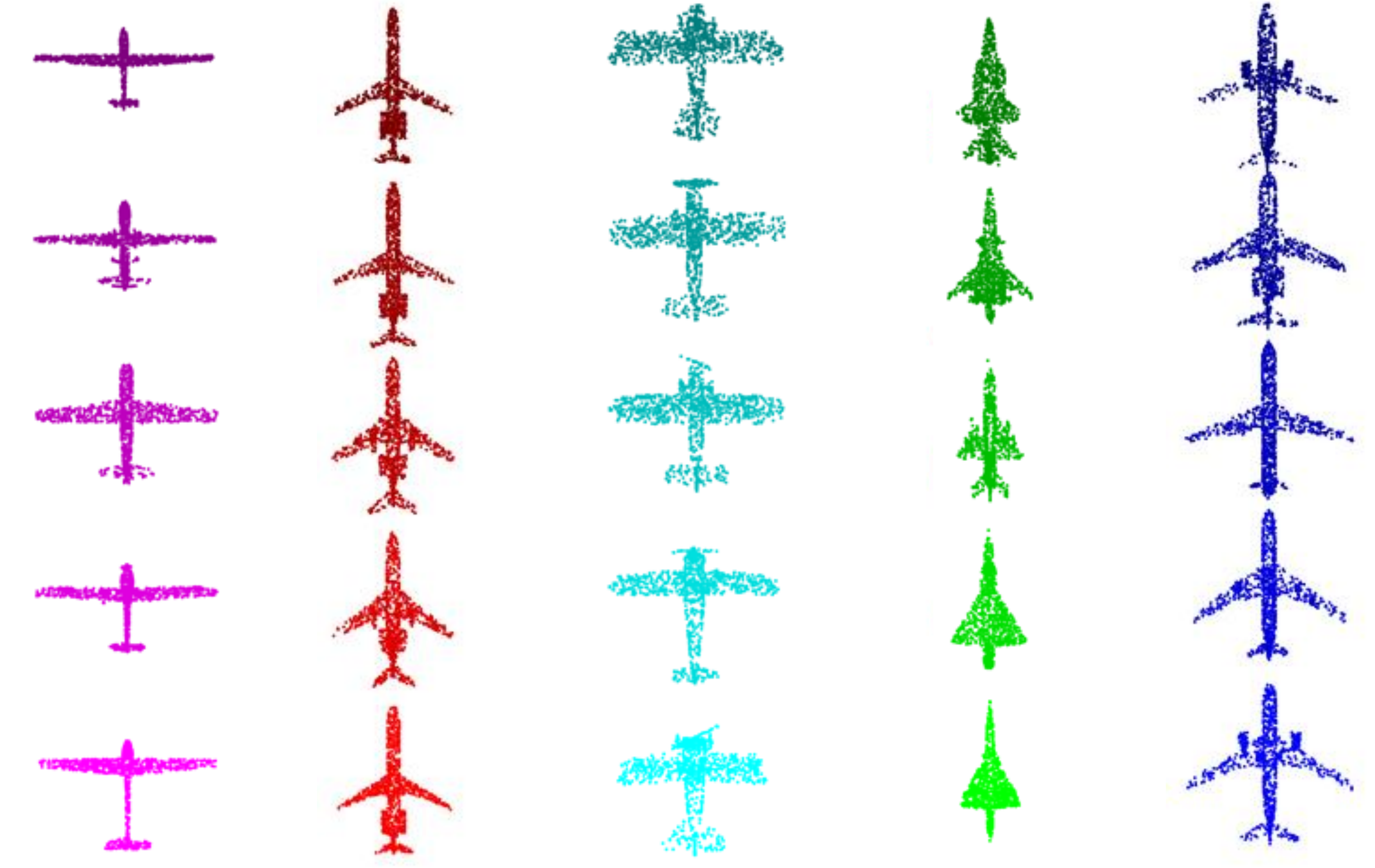}
  \caption*{Airplane}
\end{subfigure}
\vspace{0.02\textwidth}

\begin{subfigure}{.4\textwidth}
  \centering
  \includegraphics[width=.9\linewidth]{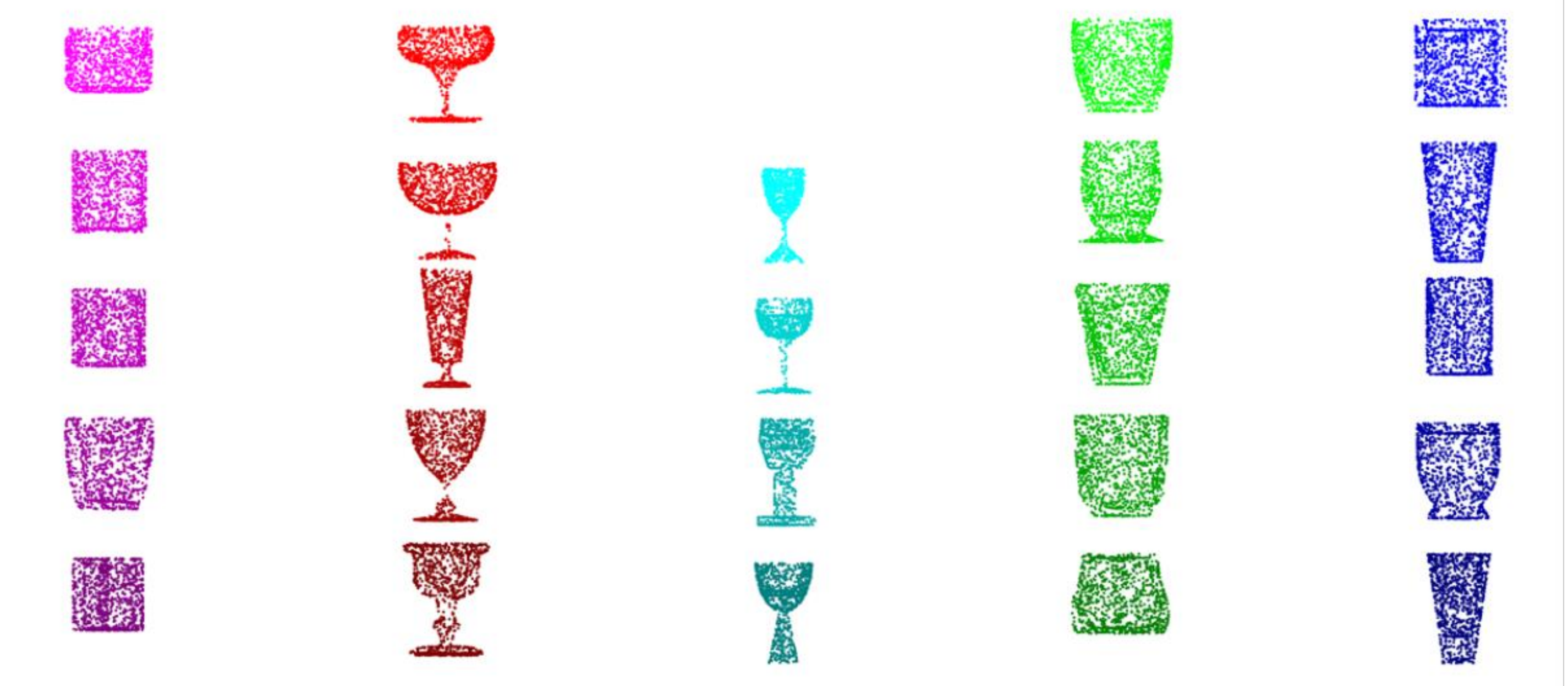}
  \caption*{Cup}
\end{subfigure}
\caption{The output of spectral clustering in the input space for the \emph{airplane} and \emph{cup} classes in the ModelNet40 dataset. \AC{The different colors within each class represent different clusters.} Our model effectively distinguishes the various subtypes of both airplanes and cups.}
    \label{fig:SC_example}
\end{figure}

Spectral clustering
\cite{citeulike:spectral,ng-spectral-clustering-analysis-2002} is
known for its ability to handle complex non-Euclidean data. To apply
it to 3D point clouds, we need to define a measure of similarity or
affinity between any pairs of point clouds. Let us restrict ourselves
to a specific class with $L$ data samples
$\mX_1, \mX_2, \ldots, \mX_{L}$ with $\mX_i \in \Real^{n_i\by 3}$. We
form an affinity matrix $\mA \in \Real^{L\by L}$ representing the
similarity between pairs of point clouds within this class, that is
$\mA_{ij} = \mathrm{affinity}(\mX_i, \mX_j)$. We will shortly discuss
how to obtain this affinity measure. Having an affinity matrix, we
form the normalized Laplacian matrix
$\mI - \mD^{-1/2} \mA \mD^{-1/2} \in \Real^{L\by L}$ and compute its
eigenvectors $\v_1, \v_2, \ldots, \v_K \in \Real^L$ corresponding to
the $K$ smallest eigenvalues, where $K$ is the intended number of
clusters. Let the rows of the matrix
$\mV_\text{input} = [\v_1, \v_2, \ldots, \v_K] \in \Real^{L \by K}$ be denoted as
$\v^1, \v^2, \ldots, \v^L \in \Real^K$. Each row $\v^i \in \Real^K$ represents the \emph{spectral embedding} of the point cloud $\mX_i \in \Real^{n_i \by 3}$ into a new space $\Real^K$. In this transformed space, the points are embedded such that affinities in the original data are preserved, making clustering more effective. The embeddings $\v^1, \v^2, \ldots, \v^L$ can then be clustered into $K$ categories using any Euclidean clustering method, such as \emph{k-means}. \cref{fig:SC_example} illustrates an application of spectral clustering on the \emph{airplane} and \emph{cup} datasets.

\subsubsection{Affinity Measure}
\label{sec:affinity}
Several methods can measure an affinity or, alternatively, a distance
between a pair of point clouds \cite{wu2021density}. Here, we select
the Chamfer Distance (CD) \cite{wu2021density} because it has a
relatively lower computational cost compared to other distance
metrics, such as the Earth Mover's Distance (EMD). Consider a pair of
point clouds $\mX_1 \in \Real^{n_1 \by 3}$ and
$\mX_2 \in \Real^{n_2 \by 3}$ and let $\x_1^i, \x_2^i \in \Real^3$ be
the $i$-th rows of $\mX_1$ and $\mX_2$, respectively. First, we register the point clouds to align together. Then, we find a mapping from each point $\x_1^i$ in
$\mX_1$ to its closest point in $\mX_2$ and vice versa:
\begin{align}
 \label{eq:closest_point}
 j^*_i = \argmin_j \|\x_1^i - \x_2^j \|,~~~ 
 i^*_j = \argmin_i \|\x_1^{i} - \x_2^j\|.
\end{align}
The Chamfer distance is then defined as the average distance of each point
in one point cloud to its nearest point in the other point cloud:
\begin{align}
\mathrm{CD}(\mX_1, \mX_2) &= \frac{1}{n_1} \sum_{i = 1}^{n_1} \|\x_1^i - \x_2^{j^*_i}\| 
\nonumber\\
&~+ 
\frac{1}{n_2} \sum_{j = 1}^{n_2} \|\x_2^j - \x_1^{i^*_j} \|.
\end{align}
Notice that, like most distance measures on point clouds, the above is
only meaningful if the two shapes are aligned together. Various
techniques exist for this, such as Iterative Closest Point (ICP)
\cite{besl1992_ICP} and Fast Point Feature Histograms (FPFH)
\cite{rusu_2009_fast}. For some datasets such as ModelNet40
\cite{wu20153d} the samples for each class are already aligned. For
datasets lacking alignment, we employ FilterReg
\cite{gao2019filterreg}, a probabilistic registration method known for
its robustness, precision, and efficiency. Notice that the alignment
is solely conducted for exemplar selection. Once
selected, the unaligned versions of the exemplars are added to the
memory. The models are 
\emph{trained and tested} on the unaligned point clouds.

It remains to form an affinity matrix from the distance
matrix. Typically a radial basis kernel is used for this
purpose. Here, we found that using the $k$-nearest neighbors
connectivity matrix for affinity improves the results. This means
that, for each sample, the affinity is set equal to 1 for its $k$
nearest neighbors and to 0 otherwise. We use $k=10$ in our
experiments. The final affinity matrix is
symmetrized as $\mA \leftarrow (\mA + \mA^T)/2$.

\subsubsection{Exemplar Selection and Memory Management}
Having segregated the data from each class into distinct clusters, we select, for every cluster, the sample nearest to its centroid in the embedded spectral space to serve as an exemplar. Within the context of continual learning, two primary memory management strategies are exercised: maintaining a fixed number of samples per class, or imposing a certain memory cap \cite{masana2022class, zhou2023deep}. Here, we opt for the first one.

\subsection{Clustering in Local Feature Space}
\label{sec:ISSFSS}
Utilizing raw input data for clustering presents advantages such as
simplicity, interpretability, efficiency, and independence from
architectural constraints. However, as we move forward in the network,
the data should have better representational properties for the specific task at hand. This motivates us to apply spectral clustering
on local features $\mZ_i \in \Real^{n_i \by F}$. We anticipate
improved results as moving closer to the feature space may result in
representatives tailored for our specific task and the specific
architecture being trained.  Similar to PointNet \cite{qi2016pointnet}
we choose $F=1024$. The main obstacle in applying spectral clustering
in such a high-dimensional space is the time and memory complexity of
the registration and computing the distance between pairs of point
clouds. Remember that for computing the Chamfer distance one needs to
align the point clouds and then for each point in one point cloud
search for the nearest neighbor in the other. To circumvent these
steps altogether, we use the nearest neighbor mappings \eqref{eq:closest_point} from the 3D
point clouds $\mX_i \in \Real^{n_i \by 3}$ to compute the Chamfer
distance in the $F$-dimensional point clouds
$\mZ_i \in \Real^{n_i \by F}$. In other words, for two sets of local
features $\mZ_1 \in \Real^{n_1 \by F}$ and
$\mZ_2 \in \Real^{n_2 \by F}$ the Chamfer distance is approximated as
\begin{align}
  \label{eq:approx_chamfer}
\tilde{\mathrm{CD}}(\mZ_1, \mZ_2) &= \frac{1}{n_1} \sum_{i = 1}^{n_1} \|\z_1^i - \z_2^{j^*_i}\|_2 
\nonumber\\
&~+ 
\frac{1}{n_2} \sum_{j = 1}^{n_2} \|\z_2^j - \z_1^{i^*_j} \|_2.
\end{align}
where the indices $j^*_i$ and $j^*_i$ are computed using
\eqref{eq:closest_point} on the corresponding 3D point clouds
$\mX_1 \in \Real^{n_1 \by 3}$ and $\mX_2 \in \Real^{n_2 \by
  3}$. Notice that, although the model has been trained with unaligned
data, to use \eqref{eq:approx_chamfer} we need to feed the network
with aligned data. Of course, when exemplars are selected, their
unaligned version is stored in the memory.

\subsection{Clustering in Global Feature Space}
\label{sec:GFS}
As discussed in \cref{sec:pointnet}, max pooling transforms the local features $\mZ_i \in \Real^{n_i \by F}$ into a single $F$-dimensional vector $\z_i \in \Real^F$, designated as the \emph{global feature}. Given their Euclidean nature, the $K$-means clustering can be straightforwardly applied to these global features. Alternatively, spectral clustering can be conducted using the Euclidean distance as a metric. Our experiments indicate that, when using global features alone, $K$-means clustering exhibits a slight advantage over spectral clustering. 

\subsection{Fusing Clustering Methods}
\label{sec:Fusion}
To further improve performance, we propose a method that integrates the previously mentioned techniques. Our core strategy involves concatenating the embedded features from each technique and then executing clustering. Remember from \cref{sec:spectral_clustering} that for a certain class with $L$ data samples $\mX_1, \mX_2, \ldots, \mX_{L}$, spectral embedding results in a matrix $\mV_\text{input} \in \Real^{L \by K}$ where $K$ is the intended number of clusters, and the $i$-th row of $\mV_\text{input}$ is the embedding of $\mX_i \in \Real^{n_i \by 3}$ into $\Real^K$. A similar matrix 
$\mV_\text{local} \in \Real^{L \by K}$ can be obtained for spectral embedding of the local features $\mZ_1, \mZ_2, \ldots, \mZ_{L}$. To fuse the two methods we simply run 
$K$-means on the rows of their horizontal concatenation $[\mV_\text{input}, \mV_\text{local}] \in \Real^{L \by 2K}$. This is straightforward due to the identical dimensions and similar scales of $\mV_\text{input}$ and $\mV_\text{local}$ stemming from their common basis in spectral analysis.

The main challenge arises when attempting to concatenate the
1024-dimensional global features
$\z_1, \z_2, \ldots, \z_{L} \in \Real^F$ with these. Our findings
suggest that direct concatenation of $\mV_\text{input}$ and $\mV_\text{local}$ with
the matrix $\mZ_\text{global} = [\z_1, \z_2, \ldots, \z_{L}]^T \in \Real^{L \by F}$
leads to a decrease in accuracy, even after experimenting with various
normalization techniques. To circumvent this, we apply spectral
embedding to the global features using the conventional Euclidean
metric to obtain $\mV_\text{global} \in \Real^{L \by K}$. Now,
$\mV_\text{global}$ matches the size and scale of the vectors $\mV_\text{input}$ and
$\mV_\text{local}$ allowing for an effective concatenation
$[\mV_\text{input}, \mV_\text{local}, \mV_\text{global}] \in \Real^{L \by 3K}$.

In \cref{sec:GFS}, we noted that the spectral embedding of global
features could degrade performance compared to applying $K$-means on
the raw features. Nevertheless, we found this approach beneficial for
merging methods. Our detailed analysis in \cref{sec:ablation}
indicates that this holistic integration of all three methods
outperforms the results of combining just two methods.

\begin{table*}[ht!]
  \setlength{\tabcolsep}{0.99\tabcolsep}

  \centering
  \tiny
    \resizebox{\textwidth}{!}{
  \begin{tabular}{@{}lcccccccccc|ccc@{}}
    \toprule
    Incremental Stage & 1 & 2 & 3 & 4 & 5 & 6 & 7 & 8 & 9 & 10 & Avg.& $\Delta$(\%)& M \\
    Number of Classes & 4 & 8 & 12 & 16 & 20 & 24 & 28 & 32 & 36 & 40 \\
    \midrule
    \textit{joint}     & 98.5 & 99.3 & 98.2 & 95.8 & 96.4 & 94.4 & 91.9 & 91.5 & 90.6 & 88.5 & 94.3 & 0 & --\\
    \textit{forgetting }& 98.5 & 54.0 & 21.3 & 21.7 & 19.9 & 20.2 & 13.2 & 11.7 & 10.7 & 9.2 & 28.0 & ↓66.3 & 0 \\
    \midrule
    I3DOL \cite{i3dol2020} & 98.1 & 97.0 & 93.4 & 91.1 & 89.7 & 88.2 & 83.5 & 77.8 & 73.1 & 61.5 & 85.3 & ↓ 9.0 & 800\\
    InOR-Net \cite{inornet2023} & 98.1 & \textbf{97.5} & \textbf{95.6} &\textbf{93.7} & 91.4 & 90.3 & 85.9 & 79.2 & 74.6 & 63.9 & 87.0 & ↓7.3 & 800\\
    \textcolor{blue}{\emph{Memory Usage\cite{i3dol2020,inornet2023}}} & \textcolor{blue}{\emph{800}} & \textcolor{blue}{\emph{800}} & \textcolor{blue}{\emph{800}} & \textcolor{blue}{\emph{800}} & \textcolor{blue}{\emph{800}} & \textcolor{blue}{\emph{800}} & \textcolor{blue}{\emph{800}} & \textcolor{blue}{\emph{800}} & \textcolor{blue}{\emph{800}}  & \textcolor{blue}{\emph{800}}  & - & - & -\\
    \midrule
    
   Ours (\textit{Input features}) & 98.8 & 95.4 & 92.6 & 91.3 & 88.7 & 86.5 & 83.2 & \underline{81.7} & \underline{79.8} & \underline{76.7} & \underline{87.5} & ↓6.8 & \textbf{400}\\

    Ours (\textit{Local features}) & 98.8 & 95.9 & 93.3 & 92.4 & 89.8 & 88.3 & 85.6 & \underline{83.8} & \underline{81.1} & \underline{78.3} & \underline{88.7} & ↓5.6 & \textbf{400}\\

    Ours (\textit{Global features}) & 98.8 & 96.2 & 93.8 & 92.7 & 90.6 & 89.7 & \underline{86.7} & \underline{85.6} & \underline{83.2} & \underline{79.1} & \underline{89.6} & ↓4.7 & \textbf{400}\\
   
    Ours (\textit{Fusion}) & 98.8 &	96.4 &	94.1 &	93.2	& \textbf{91.5} &	\textbf{90.7}	&\textbf{87.1}&	\textbf{86.3}	&\textbf{85.1}	& \textbf{80.8}&	\textbf{90.4} & ↓\textbf{3.9}& \textbf{400} \\

     \textcolor{blue}{\emph{Memory Usage (Ours)}} & \textcolor{blue}{\emph{40}} & \textcolor{blue}{\emph{80}} & \textcolor{blue}{\emph{120}} & \textcolor{blue}{\emph{160}} & \textcolor{blue}{\emph{200}} & \textcolor{blue}{\emph{240}} & \textcolor{blue}{\emph{280}} & \textcolor{blue}{\emph{320}} & \textcolor{blue}{\emph{360}}  & \textcolor{blue}{\emph{400}}  & - & - & -\\
    
    \bottomrule
  \end{tabular}
}

\caption{ Accuracy comparison on ModelNet40 Dataset\cite{wu20153d}
  over 10 incremental stages. The final columns indicate the average
  accuracy across all stages (Avg.), the forgetting rate
  (\(\Delta\%\)), and the total number of exemplars stored in memory
  (M).  We also report the number of exemplars in memory (\textcolor{blue}{\textit{Memory Usage}}) for each stage. \AC{Additionally, "\textit{joint}" refers to learning all classes together, while "\textit{forgetting}" learns stages consecutively without preventing forgetting.} The best results are highlighted in
  \textbf{bold}, and our improvements are
  \underline{underlined}. Notice that the slightly lower performance
  of our approach in the initial stages is a result of significantly
  reduced memory usage, for instance, only utilizing one-tenth and
  one-fifth of InOR-net's memory in stages 2 and 4, respectively.  }
  \label{tab:ModelNet}
\end{table*}

\subsection{Class Imbalance Problem}
\label{sec:CIP}
A principal challenge in memory-based continual learning scenarios is the substantial imbalance between the exemplars of the previously learned classes and the samples of the newly introduced classes. Several strategies have been proposed to address this issue. For example, \cite{GeoDL_2021_CVPR} utilizes a cosine distance metric to mitigate bias in the output layer, while \cite{BiC_2019_CVPR} involves learning a bias-correction model for post-processing output logits. In this work, we explore the use of \emph{focal Loss} \cite{Focal_2017_ICCV}, initially introduced for object detection tasks. Our experiments indicate that this approach can serve as a novel yet effective solution in the context of continual learning. The  focal loss \cite{Focal_2017_ICCV} can be defined as
\begin{equation}
\mathrm{FL}(p_t) = -\alpha_t \,(1 - p_t)^\gamma \,\log(p_t),
\end{equation}
where $p_t$ is the model's estimated probability for each class, $\alpha_t$ is a balancing parameter to address class imbalance, and $\gamma$ is a focusing parameter to steer the model's focus towards hard examples. By focusing on the harder examples from these new classes, the model can learn more effectively from limited data.

\section{Experiments}
\label{sec:formatting}

\subsection{Settings}

\AC{We follow the experimental setup from I3DOL \cite{i3dol2020} and InOR-net \cite{inornet2023}, including the datasets, backbone, and number of incremental stages. To the best of our knowledge, these are the only works focused on exemplar-based continual learning for point clouds. Non-exemplar-based methods have reported significantly lower accuracy \cite{PointCLIMB_2023}.}

\paragraph{Datasets.}  We conduct tests on three datasets:
ModelNet40 \cite{wu20153d}, ShapeNet \cite{chang2015shapenet}, and
ScanNet \cite{dai2017scannet}. ModelNet40 \cite{wu20153d} contains 40 distinct classes of clean 3D
CAD models.  Following \cite{i3dol2020} and \cite{inornet2023}, we
establish a total of 10 incremental stages with every stage adding four
new classes. We store 10 exemplars from each class, leading to an
overall memory usage of 400 samples.

For the ShapeNet Dataset \cite{chang2015shapenet} we use 53 categories
to be consistent with I3DOL \cite{i3dol2020} and InOR-net \cite{inornet2023}.
We follow their setting of 9 incremental stages with each stage
introducing six new classes, except for the final stage, which introduces five classes. There is no need for sample alignment in our
procedures, as the ShapeNet dataset \cite{chang2015shapenet} already
arrives in an aligned format.

The \emph{ScanNet} dataset includes 17 different classes, all obtained
from scanning real indoor scenes. Compared to ShapeNet
\cite{chang2015shapenet} and ModelNet40 \cite{wu20153d},
ScanNet \cite{dai2017scannet} presents a greater challenge for our
framework due to its noisy geometric structures and lack of
alignment. We apply the FilterReg \cite{gao2019filterreg} technique
for aligning samples to compute a reliable distance matrix. Our model
stores 10 samples per class, with a total of 9 incremental states each
adding two new classes, apart from the final state adding just one.
\\

 \begin{table*}[ht!]
  \setlength{\tabcolsep}{0.9\tabcolsep}

  \centering
  \tiny
    \resizebox{\textwidth}{!}{
  \begin{tabular}{@{}lccccccccc|ccc@{}}
    \toprule
    Incremental Stage & 1 & 2 & 3 & 4 & 5 & 6 & 7 & 8 & 9 & Avg.& $\Delta$(\%)& M \\
    Number of Classes & 6 & 12 & 18 & 24 & 30 & 36 & 42 & 48 & 53 \\
    \midrule
    \textit{joint} & 98.0 & 97.3 & 95.4 & 94.4 & 92.8 & 91.6 & 90.9 & 90.5 & 89.3 & 93.3 & 0 & --\\
    \textit{forgetting} & 98.0 & 46.3 & 53.4 & 4.5 & 11.8 & 6.6 & 3.2 & 12.5 & 6.2 & 26.9 & ↓66.4 & 0 \\

    \midrule
    I3DOL \cite{i3dol2020} & 97.5 & 94.4 & 90.2 & 84.3 & 80.5 & 76.1 & 73.5 & 70.8 & 67.3 & 81.6 & ↓11.7 & 1000\\
    InOR-Net \cite{inornet2023}  & 97.5 &\textbf{95.6}  & \textbf{92.4} & 86.7 & 83.1 & 79.2 & 76.0 & 73.5 & 69.4 & 83.7 & ↓9.6 & 1000\\

    \textcolor{blue}{\emph{Memory Usage \cite{i3dol2020,inornet2023}}} & \textcolor{blue}{\emph{1000}} & \textcolor{blue}{\emph{1000}} & \textcolor{blue}{\emph{1000}} & \textcolor{blue}{\emph{1000}} & \textcolor{blue}{\emph{1000}} & \textcolor{blue}{\emph{1000}} & \textcolor{blue}{\emph{1000}} & \textcolor{blue}{\emph{1000}} & \textcolor{blue}{\emph{1000}}  & - & - & -\\
    
    \midrule
    
    Ours (\textit{Input features})   & 98.0 & 91.7 & 90.2 & \underline{86.9} & \underline{84.2} & \underline{80.1} & \underline{77.4} & \underline{74.1} & \underline{69.6} & 83.5 & ↓9.8 & \textbf{530}\\
    
    Ours (\textit{Local features})   & 98.0 & 91.6 & 90.3 & \underline{87.2} & \underline{84.9} & \underline{81.4} & \underline{78.3} & \underline{76.1} & \underline{70.3} & \underline{84.2} & ↓9.1 & \textbf{530}\\
    
    Ours (\textit{Global features})   & 98.0 & 91.8 & 90.9 & \underline{87.8} & \underline{85.7} & \underline{82.6} & \underline{80.2} & \underline{77.5} & \underline{71.3} & \underline{85.1} & ↓8.2 & \textbf{530}\\
    
     Ours (\textit{Fusion})  &  98.0 & 91.8 & 90.8 & \textbf{88.2} & \textbf{86.5} & \textbf{83.4} & \textbf{81.5} & \textbf{78.6} & \textbf{72.4} & \textbf{85.7} & ↓\textbf{7.6} & \textbf{530} \\

    \textcolor{blue}{\emph{Memory Usage (Ours)}} & \textcolor{blue}{\emph{60}} & \textcolor{blue}{\emph{120}} & \textcolor{blue}{\emph{180}} & \textcolor{blue}{\emph{240}} & \textcolor{blue}{\emph{300}} & \textcolor{blue}{\emph{360}} & \textcolor{blue}{\emph{420}} & \textcolor{blue}{\emph{480}} & \textcolor{blue}{\emph{530}}  & - & - & -\\

    \bottomrule
  \end{tabular}
}
  \caption{Accuracy comparison on ShapeNet  \cite{chang2015shapenet} over 9 incremental stages. Similar to \cref{tab:ModelNet}, lower performance on stages 1 and 2 is due to significantly smaller memory usage compared to InOR-net.}
  \label{tab:ShapeNet}
\end{table*}


\paragraph{Implementation Details.} CL3D is implemented with PyTorch
and trained on a Tesla V4 GPU with a batch size of 32. To maintain a
fair comparison with previous studies \cite{i3dol2020,inornet2023}, we
use PointNet \cite{qi2016pointnet} as our backbone.  We choose the
Adam optimizer and the cosine annealing learning rate schedule with an
initial learning rate of $10^{-3}$. The number of epochs is set to 50
in incremental stages. We set the distillation factor to 0.1 for the
knowledge distillation loss and choose $\gamma = 2$ for the focal loss.

\subsection{Comparison}
We conduct comparative analyses with I3DOL \cite{i3dol2020} and
InOR-net \cite{inornet2023}. \AC{For a comprehensive perspective, we
report the scenario where the network has access to the complete dataset
from previous tasks (joint training) as an ideal upper bound. As a
lower bound, we present the results of full forgetting (denoted as \textit{forgetting} in the tables) where the model updates its parameters solely based on new tasks.}

\paragraph{Results on ModelNet40.}
\cref{tab:ModelNet} demonstrates a comparison on the ModelNet40
dataset\cite{wu20153d}. By integrating the spectral embeddings from
input, local, and global features, our model achieves a 3.4\% increase
in average accuracy compared to InOR-net \cite{inornet2023}, while
using substantially less memory in every stage. In the final stage, we
achieve a remarkable 16.9\% increase in accuracy compared to InOR-net,
using only half the memory. It is notable that just using the input
space features, we perform on par with the state of the art, enhancing
average accuracy by 0.5\%. 
We reiterate that using solely the input data removes any reliance on the
network architecture. Additionally, note that the average accuracy also
takes into account the initial stages where we use considerably less memory. 
The forgetting rate has decreased to 6.8\% for
the input approach and to 3.9\% for the fused features.

\paragraph{Results on ShapeNet.}
\cref{tab:ShapeNet} shows our result on the ShapeNet dataset
\cite{chang2015shapenet}, which has more classes than
ModelNet\cite{wu20153d} and ScanNet\cite{dai2017scannet}. On this
dataset, our fusion method achieved an accuracy increase of 2.0\%
compared to InOR-net \cite{inornet2023}, while using only 10 samples
per class in memory. In the final stage, we beat InOR-net by 3.0\%
while using nearly half the memory. Our forgetting rate is 9.8\% for
input features and 7.6\% for the fused features.

\begin{table*}[thb!]
  \setlength{\tabcolsep}{0.99\tabcolsep}
  \centering
  \tiny
    \resizebox{\textwidth}{!}{
  \begin{tabular}{@{}lccccccccc|ccc@{}}
    \toprule
    Incremental Stage & 1 & 2 & 3 & 4 & 5 & 6 & 7 & 8 & 9 & Avg.& $\Delta$(\%)& M \\
    Number of Classes & 2 & 4 & 6 & 8 & 10 & 12 & 14 & 16 & 17 \\
    \midrule
    \textit{joint}      & 96.8 &	94.8 & 93.4 & 93.2 & 92.9 &	92.3 & 91.8 & 91.3 & 91.0 & 93.0 & 0 & --\\
    \textit{forgetting} & 95.6 & 49.8 & 32.5 & 23.7 & 20.2 & 13.8 &	15.0 & 12.3 & 8.2 &	30.1 & ↓66.3 & 0 \\
    
    \midrule
    I3DOL \cite{i3dol2020} & 93.2 & 87.2 & 80.5 & 77.8 & 64.3 & 61.9 & 58.2 & 56.8 & 52.1 & 70.2 & ↓22.8 & 600\\
    InOR-Net \cite{inornet2023}  & 93.2 & \textbf{88.7} & \textbf{82.6} & \textbf{79.4} & 67.9 & 64.0 & 60.6 & 58.3 & 54.8 & 72.2 & ↓20.8 & 600\\
        \textcolor{blue}{\emph{Memory Usage\cite{i3dol2020,inornet2023}}} & \textcolor{blue}{\emph{600}} & \textcolor{blue}{\emph{600}} & \textcolor{blue}{\emph{600}} & \textcolor{blue}{\emph{600}} & \textcolor{blue}{\emph{600}} & \textcolor{blue}{\emph{600}} & \textcolor{blue}{\emph{600}} & \textcolor{blue}{\emph{600}} & \textcolor{blue}{\emph{600}}  & - & - & -\\
    \midrule
    
    Ours (\textit{Input features})  & 97.5 &	79.4 &	75.5 &	74.3 &	\underline{73.1} &	\underline{71.7} &	\underline{70.2} &	\underline{67.2} &	\underline{63.4} & \underline{74.6} & ↓18.4 & \textbf{170}\\
    
    Ours (\textit{Local features})  & 97.5 & 78.7 & 75.2 & 74.5 & \underline{73.4} & \underline{71.3}	& \underline{69.5} & \underline{66.3} & \underline{62.2} & \underline{74.3} & ↓18.7 & \textbf{170} \\
    
    Ours (\textit{Global features})   & 97.5 & 79.3 & 75.8 & 75.1 & \underline{74.2} & \textbf{73.3}	& \underline{71.5} & \underline{69.6} & \underline{65.1} & \underline{75.7} & ↓17.3 & \textbf{170} \\

    Ours (\textit{Fusion})  & 97.5 & 78.8 & 75.6 & 75.3 & \textbf{74.5} & \underline{73.1}	& \textbf{72.2} & \textbf{71.4} & \textbf{68.8} & \textbf{76.3} & ↓16.7 & \textbf{170} \\

    \textcolor{blue}{\emph{Memory Usage (Ours)}} & \textcolor{blue}{\emph{20}} & \textcolor{blue}{\emph{40}} & \textcolor{blue}{\emph{60}} & \textcolor{blue}{\emph{80}} & \textcolor{blue}{\emph{100}} & \textcolor{blue}{\emph{120}} & \textcolor{blue}{\emph{140}} & \textcolor{blue}{\emph{160}} & \textcolor{blue}{\emph{170}}  & - & - & -\\
    
    \bottomrule
  \end{tabular}
}
  \caption{Performance comparison on ScanNet dataset \cite{dai2017scannet} over 9 incremental stages. As in \cref{tab:ModelNet} and \cref{tab:ShapeNet}, the lower performance in the initial stages is due to significantly reduced memory usage compared to InOR-net utilizing just 7\%, 10\%, and 14\% for stages 2, 3, and 4, respectively.}
\end{table*}

\paragraph{Results on ScanNet.}
The ScanNet dataset \cite{dai2017scannet}, as a real-world point cloud dataset, presents increased challenges due to its noisiness, incompleteness, and other complex characteristics. By utilizing 10 samples per class in our fusion approach, we achieve 
an increase of 4.1\% in average accuracy over the state of the art. In the final stage, we obtain a 14.0\% increase in accuracy while 
using only 28\% of the memory used by InOR-net \cite{inornet2023}. Our results illustrate that in real-world scenarios, proper selection of exemplars can make a significant difference and greatly affect learning in the next stages. The imperfect quality of point clouds increases the risk of selecting suboptimal samples. 


\subsection{Ablation Study}
\label{sec:ablation}
\begin{figure}[h!]
  \centering
   \includegraphics[width=0.9\linewidth]{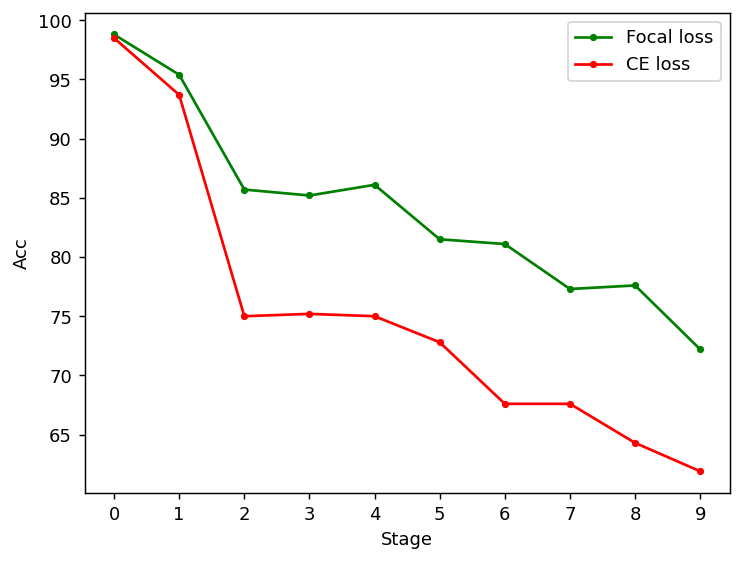}

   \caption{The effect of focal loss compared to Cross-Entropy loss on classification accuracy in a continual learning setting using input space spectral features with 5 samples per class on ModelNet40 \cite{wu20153d} }
   \label{fig:FvsL}
\end{figure}

\paragraph{The Effect of Focal Loss.} As discussed in \cref{sec:CIP}, we have implemented \emph{focal loss} \cite{Focal_2017_ICCV} to address class imbalance. \cref{fig:FvsL} clearly demonstrates the effectiveness of focal loss when using only the input space spectral features with 5 samples per class. The 
comparison with the \emph{Cross-Entropy loss} highlights the significant impact of focal loss on maintaining model accuracy through the various stages of continual learning.

\begin{figure}[h]
  \centering
       \includegraphics[width=0.8\linewidth]{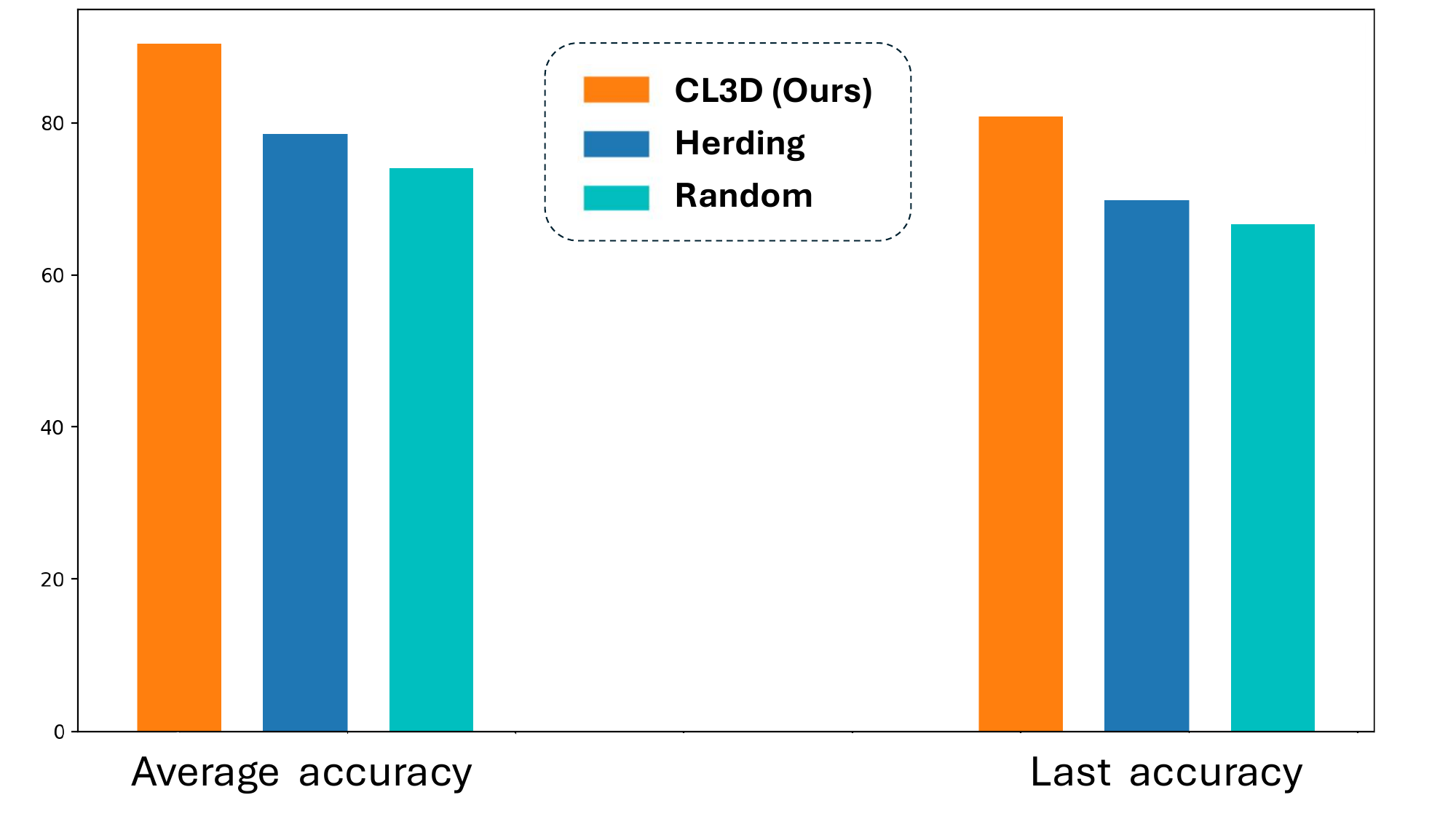}
       \vspace{-5pt}
    \caption{\AC{Comparing clustering-based exemplar selection with \emph{herding} and random selection, reporting the \emph{average accuracy} (left) and the accuracy in the last stage (right) on ModelNet \cite{wu20153d}.}}
    \label{fig:Herding}
\end{figure}

\vspace{-0.5em}
\paragraph{Replace Clustering with Herding or Random Selection.} To assess whether our approach's success is due to clustering or other factors like distillation or focal loss, we replaced the clustering-based exemplar selection with herding \cite{Rebuffi_2017_CVPR} or random selection, keeping all other components unchanged. The results, depicted in \cref{fig:Herding}, clearly indicate that clustering indeed plays a significant role in boosting our method's performance.

\vspace{-20pt}

\paragraph{Feature Fusion.} 
It is natural to question whether incorporating all three types of features—input, local, and global—truly enhances the overall accuracy. \cref{tab:fusion} presents accuracy metrics for various combinations of these features. The data clearly indicate that combining input, local, and global features leads to superior performance compared to utilizing them individually or in pairs.

\setlength{\tabcolsep}{10pt} 
\begin{table}[ht]
\centering
\begin{tabular}{cccccc}
\toprule
Input & Local & Global & $ACC_{avg}$ & $ACC_{last}$ \\
\midrule
$\checkmark$ &  &  & 87.8  & 76.7  \\
$\checkmark$ & $\checkmark$ &  & 88.5  & 77.4  \\
 & $\checkmark$ &  & 88.7  & 78.3  \\
 $\checkmark$ &  & $\checkmark$ & 88.8  & 78.8  \\
 &  & $\checkmark$ & 89.6  & 79.1  \\
 & $\checkmark$ & $\checkmark$ & 89.7  & 79.5  \\
$\checkmark$ & $\checkmark$ & $\checkmark$ & {\bf 90.4}  & \textbf{80.8}  \\

\bottomrule
\end{tabular}
  \caption{Accuracy comparison on ModelNet40 \cite{wu20153d} with 10 samples per class, using different combinations of input, local, and global features. The table shows the average accuracy ($ACC_{avg}$) and the accuracy at the last stage ($ACC_{last}$). Combining all three features (Input, Local, Global) achieves the highest average accuracy (90.4\%) and last accuracy (80.8\%)}
  \label{tab:fusion}
\end{table}   

\section{Discussion} 
\label{sec:discussion}

Our CL3D model demonstrates strong performance but faces challenges. The current exemplar selection method, based on choosing $K$ clusters, results in a linear increase in memory usage, limiting our ability to maintain a fixed memory budget. A transition to hierarchical clustering could allow for a more adaptive exemplar selection across stages. \AC{Additionally, the most computationally expensive step is computing the affinity between all point cloud samples within a class, leading to quadratic complexity. Fortunately, this issue has been addressed extensively in the context of spectral clustering \cite{huang2019ultra,boutsidis2009random,charless2004spectral}.}

\section{Conclusion}
In this paper, we introduce CL3D, a novel framework for continual learning in 3D point cloud objects. By leveraging spectral clustering in input, local, and global feature spaces, we effectively identify key exemplars for continual learning. Our approach is backbone-independent in the input section, making it adaptable for future methods. Extensive experiments on ModelNet40, ShapeNet, and ScanNet demonstrate that CL3D achieves superior accuracy with reduced memory requirements compared to existing methods.


{\small
\bibliographystyle{ieee_fullname}
\bibliography{Ref}
}

\end{document}